\documentclass[conference]{IEEEtran}
\IEEEoverridecommandlockouts
\usepackage{cite}
\usepackage{amsmath,amssymb,amsfonts}
\usepackage{graphicx}
\usepackage{textcomp}
\usepackage{xcolor}
\def\BibTeX{{\rm B\kern-.05em{\sc i\kern-.025em b}\kern-.08em
    T\kern-.1667em\lower.7ex\hbox{E}\kern-.125emX}}

\usepackage{comment}

\usepackage{booktabs} 
\usepackage{tabularx} 
\usepackage{multirow} 
\usepackage{balance}
\usepackage{xargs}                      

\usepackage{algorithm}
\usepackage[noend]{algpseudocode}
\usepackage{csquotes}

\usepackage[figureposition=bottom,tableposition=bottom,font=footnotesize]{caption} 
\algnewcommand{\LineComment}[1]{\State \(\triangleright\) #1}

\begin{document}

\title{Efficient Parking Search using Shared Fleet Data}

\author{\IEEEauthorblockN{Niklas Strau\ss}
\IEEEauthorblockA{\textit{Institute for Informatics} \\
\textit{LMU Munich}\\
Munich, Germany \\
strauss@dbs.ifi.lmu.de}
\and
\IEEEauthorblockN{Lukas Rottkamp}
\IEEEauthorblockA{\textit{Institute for Informatics} \\
\textit{LMU Munich}\\
Munich, Germany \\
lukas.rottkamp@campus.lmu.de}
\and
\IEEEauthorblockN{Sebastian Schmoll}
\IEEEauthorblockA{\textit{Institute for Informatics} \\
\textit{LMU Munich}\\
Munich, Germany \\
schmoll@dbs.ifi.lmu.de}
\and
\IEEEauthorblockN{Matthias Schubert}
\IEEEauthorblockA{\textit{Institute for Informatics} \\
\textit{LMU Munich}\\
Munich, Germany \\
schubert@dbs.ifi.lmu.de}
}

\maketitle

\begin{abstract}
Finding an available on-street parking spot is a relevant problem of day-to-day life. In recent years, cities such as Melbourne and San Francisco deployed sensors that provide real-time information about the occupation of parking spots. Finding a free parking spot in such a smart environment can be modeled and solved as a Markov decision process (MDP). The problem has to consider uncertainty as available parking spots might not remain available until arrival due to other vehicles also claiming spots in the meantime. Knowing the parking intention of every vehicle in the environment would eliminate this uncertainty. Unfortunately, it does currently not seem realistic to have such data from all vehicles. In contrast, acquiring data from a subset of vehicles or a vehicle fleet appears feasible and has the potential to reduce uncertainty. 

In this paper, we examine the question of how useful sharing data within a vehicle fleet might be for the search times of particular drivers. We use fleet data to better estimate the availability of parking spots at arrival.  Since optimal solutions for large scenarios are infeasible, we base our method on approximate solutions, which have been shown to perform well in single-agent settings. Our experiments are conducted on a simulation using real-world and synthetic data from the city of Melbourne. The results indicate that fleet data can significantly reduce search times for an available parking spot.
\end{abstract}

\begin{IEEEkeywords}
Parking, Multi-Agent Routing, Sequential Decision Making Under Uncertainty 
\end{IEEEkeywords}

\section{Introduction}
\label{section:introduction}
Finding resources like on-street parking spots or charging stations in an urban environment can be very annoying and time-consuming. 
According to \cite{Shoup2006}, on average 30\% of traffic in cities is caused by cars cruising to find a parking spot. Minimizing the time spent to find a parking spot has many beneficial effects on the environment, the stress level of drivers and traffic in cities.

In recent years, several cities including San Francisco \cite{Gauthier2012} and Melbourne \cite{Liu2018} deployed systems that provide real-time information about the occupation of parking spots through in-ground sensors. Consequently, a number of smart parking systems emerged \cite{Al-Turjman2019}.
Knowing the current occupation state of a parking spot does not eliminate uncertainty about its availability upon arrival as other drivers may claim that spot in the meantime or a parking spot closer to the driver's destination might become available. Guiding an agent to a free parking spot in such an environment is an instance of the resource routing problem, a probabilistic search problem for routing users to a resource which is available upon arrival. Another example of resource routing is the search for a charging station for a battery-electric vehicle. The authors of \cite{Schmoll2018} formulate the problem of finding a vacant resource for a single driver in such a smart environment as a Markov decision process (MDP). This approach allows combining information about the current availability of parking spots with a statistic model about future resource states that has been trained on past observations.
In such a single-agent setting, all other drivers are not modeled explicitly. Instead, their impact on the environment is modeled by the probabilistic process describing future resource states.

In real-world scenarios, several drivers may want to reach destinations in the same area at the same time and thus compete over vacant parking spots. Having knowledge about the parking intentions of all vehicles would allow to eliminate uncertainty about future resource states. Unfortunately, it currently seems unlikely that all vehicles share their data on a common platform in the near future. 
However, a subset of vehicles using the same parking route guidance system is realistic, e.g. cars by a certain manufacturer or rental company. These can share their positions and destinations. Furthermore, since they use the same system, we can assume to also have a good prediction of future routing decisions. We will refer to such a subset of vehicles as vehicle fleet and examine the question of whether the shared information can be used to reduce the average search times of fleet vehicles.
In particular, we propose different approaches that incorporate vehicle fleet information in order to reduce uncertainty about future resource states and evaluate their use to improve parking guidance systems. Since optimal solutions are infeasible for larger scenarios, we base our approaches on two approximate solvers, replanning and hindsight planning. These have been shown to perform well in single-agent settings \cite{Schmoll:2019:SDR:3340964.3340983}. We model fleet parking guidance as a competitive environment, where drivers act selfishly.
In other words, every driver only tries to minimize his or her own search time regardless of the influence on other fleet vehicles.

Our first approach is called \enquote{reservations} and reduces the aforementioned uncertainty through agents sharing their currently targeted parking bay. The reasoning is that any other fleet vehicle heading for the same parking spot and arriving earlier than the agent dramatically reduces the probability that the spot is available when the agent arrives. Reservations allow agents to consider such parking spots as occupied even if they are currently not.
However, reservations cannot capture the uncertainty regarding an agent's parking intention, which arises from vehicles outside the fleet. Our second approach, multi-agent dynamic probability adaption, tries to account for that by also adjusting the availability probability of nearby resources through a heuristic that approximates the behavior of agents in case their intended parking spot becomes occupied.

We evaluate the proposed approaches through agent-based simulations, using real-world occupancy data from Melbourne. We also facilitate synthetic data to analyze a broader variety of situations. 
The evaluation shows that our proposed multi-agent improvements reduce the time spent to find a parking spot by up to around 84\%. 

To summarize, the contributions of this paper are:
\begin{itemize}
    \item Formalizing resource routing in fleet scenarios.
    \item Fleet-based resource search based on reservations.
    \item Advanced methods for fleet-based resource search using adaption heuristics.
\end{itemize}

This paper is structured as follows: Section~\ref{rel_work} provides an overview of related work. In section~\ref{section:methodology}, we define the problem setting and describe the single-agent solutions our methods are based on. Our novel solutions for fleet scenarios in the presence of real-time data from stationary sensors are presented in sections~\ref{sec:reservations} and~\ref{sec:prob_adaption}. We then continue by evaluating all proposed algorithms in an agent-based simulation in section~\ref{section:simulator}. Finally, we conclude the paper in section~\ref{sec:conclusion}.

\section{Related Work}

\label{rel_work}
Searching for an on-street parking spot is a very common problem relevant to everyday life. 
Consequently, a vast number of approaches have been proposed to either predict the availability of parking spots, simulate the behavior of agents looking for parking spots or find routes that minimize the time searching for a parking spot \cite{Al-Turjman2019}.
Many existing parking guidance approaches do not solve competition between users, rely on hardware to limit access to parking spots or assume that all drivers are guided by the same system.
\label{sec:parkagent}
The authors of \cite{Benenson2008} propose an agent-based simulation that tries to replicate the behavior of a human driver. The decision where to park is based upon observations about the vacancy of parking spots around the destination.
\cite{Basu2002} propose parking meters which report their availability in real-time to a system so that users can navigate to available parking spots. They introduce a variation of their approach: In cases when two agents compete over a single resource, it will only be seen as available for the driver which is closer to that parking spot. This approach is very similar to the concept of reservations we will present later in this paper. However, their system does not consider the chance that a parking spot might be occupied by third vehicles which is not part of the fleet.

The authors of \cite{Liu2018} introduce a model to predict the utilization rate of a street segment which combines historical as well as real-time data. At each decision point, the system recommends one street segment out of a set of candidates matching the user's preferences.
They also extend their approach to multi-user scenarios: A multi-user factor is applied to decrease the occupancy probabilities when other agents arrive at a street segment earlier.
This approach shows some similarity to our multi-agent probability adaption described below. It also relies on modifying the availability probabilities of resources.
As they rely on Integer Programming to minimize the objective function, their approach is NP-hard \cite{Karp2010} and may be difficult to apply in an online manner when a huge number of agents and resources is present.
\cite{Geng2011} assign users to parking spots through a central allocation system by using a queuing model. A major drawback of their implementation is that it relies on hardware to make parking spots only accessible to users with a reservation.

The authors of  \cite{Josse2013} try to find routes that maximize the likelihood of getting a parking spot by adopting the Traveling Salesmen problem.
\cite{Verroios2011} are computing routes to available parking spots through solving a time-varying traveling salesmen problem. In their approach, agents observe the state of resources they pass by and communicate it to other agents in the area. In contrast to the approaches in this paper, they ignore resources which are occupied and may become available during the search.
The authors of \cite{Ayala2011} use a game-theoretic framework for the search of a parking spot in competitive multi-agent settings. The goal of such a game is to find an assignment of agents to parking spots that minimizes the costs of the assignment.
Their setting differs from our setting as they assume that either all vehicles are part of the system or information about other vehicles is only estimated by a prior probability distribution.

\cite{Schmoll2018a} define parking search using sensor data as an instance of the dynamic resource routing (DRR) problem and solve it as a fully observable Markov decision process (MDP). A policy is computed by bounded real-time dynamic programming using novel bounds and estimation methods. Even though the authors considerably lowered the computational overhead compared to other MDP solvers, the exponential growth of the state space still limits the applicability to larger settings. Thus, \cite{Schmoll:2019:SDR:3340964.3340983} propose novel approximate methods based on replanning and hindsight planning. The results indicate that search times come close to the optimal solution of \cite{Schmoll2018a} if these can be determined and provide efficient policies for large settings.
Unlike our approaches, their work does not consider explicit information about other vehicles searching for parking spots.
Our approaches are based on \cite{Schmoll:2019:SDR:3340964.3340983} and we will include their original methods as a single-agent baseline to determine the benefits of our proposed fleet solutions.

\cite{Bock2019} have shown that the ability to predict the availability of a parking spot in the near future is crucial for the performance of dynamic resource routing. There exists a vast number of approaches to predict the availability of parking spots \cite{Rottkamp2018,Shao2019,Jomaa2019,Zheng2015,Arjona2019, Vlahogianni2016, Bock2017}. However, predicting the expected occupancy of a group of resources is not equivalent to predicting the probability distribution that a parking spot is vacant at a certain time, which is necessary for many approaches.
In this paper, we focus on the use of a continuous-time Markov chain (CTMC) as proposed in \cite{Josse2015, Schmoll:2019:SDR:3340964.3340983} to incorporate recent observations (e.g. current state of a resource) as well as long term observations (e.g. average occupancy time).

\section{Methodology}
\label{section:methodology}
In this section, we formalize the search for a free parking spot in a fully observable multi-agent setting as a Markov decision process and recapitulate the approximate solutions \cite{Schmoll:2019:SDR:3340964.3340983} our methods are based on. 

\subsection{Problem Setting}
The goal of DRR (dynamic resource routing) is to guide an agent to a resource $r_i \in R$, which is available upon arrival, in a directed graph $G = (N,E,C)$, such that the expected total travel time is minimized. The graph $G$ represents a road network: Nodes $N$ correspond to intersections, edges $E = N \times N$ to road segments and $C: E \rightarrow R^+$ is a function that defines the cost of traversing an edge. The cost of an edge can be estimated by the distance and the speed limit. Resources represent parking spots and are associated with an edge in the graph. Each driver or agent $\alpha_i$, from the set of agents $\Lambda = \{ \alpha_i\}$, has a different final destination, start intersection and can begin the trip from start to destination at any time. As this information is not known in advance, the set of agents $\Lambda$ is not stationary over time. We define $c_t(r,\alpha_i)$ to be the terminal cost of a resource and an agent. In case of parking spots, this is the time for walking from the parking spot to the final destination. The total travel time consists of the driving time and the terminal costs $c_t$. Every time an agent is at an intersection, a decision needs to be made whether to continue driving to another intersection or to park at a specific spot located on an outgoing edge of the intersection. We call a setting fully observable if each agent always has information about the current state of all resources in $R$.

In our setting, multiple drivers are simultaneously looking for a parking spot in the same area. This leads to competition over parking spots which can be reduced by adjusting the search policy. Even though agents act selfishly, reducing conflicts still benefits them because heading for a spot being most likely occupied by another fleet vehicle can be avoided. In a competitive and independent multi-agent setting, a separate MDP is associated with each agent $\alpha$. An MDP can be defined as a 4-tuple $(S, A, C, P)$, where $S$ is the (possibly infinite) set of all possible states and $A$ denotes the set of actions. Furthermore, $A_s$ represents the set of available actions while in state $s \in S$. $C: A \rightarrow R$ is the function that defines the cost for an action $a \in A$. $P: A \times S \rightarrow [0,1]$ describes the probability for traversing from state $s \in S$ to state $s' \in S$ after choosing action $a \in A_s$.

A policy $\pi(s) = a$ is a mapping from any state $s\in S$ to an action $a \in A_s$. Solving an MDP corresponds to finding a policy that minimizes the expected future costs over an infinite time horizon. In our setting, we assume that all fleet vehicles are guided by the same system and thus, policies vary only with respect to the agent's destination. Therefore, for a given agent policy, the policies of all other agents can be considered as static.
The expected future costs are commonly denoted as utility $U$ with the following bellman equation:
\begin{equation}
U^\pi(s) = C(s_t, \pi(s_t)) + \sum_{s' \in S} P(s' | \pi(s), s) U^\pi(s')
\end{equation}
The Q-value of a state-action pair $(a, s)$, where  $a \in A_s$ and $s \in S$, describes the expected costs when performing the action $a$ and following the optimal policy $\pi$. It is defined as follows:
\begin{equation} \label{eq:q_func}
Q_\pi(s,a) = C(s,a ) + \sum_{s' \in S} P(s' | a, s) U^{\pi}(s') 
\end{equation}
An optimal policy consists of taking the actions with the lowest Q-Value:
\begin{equation} \label{eq:optimal_policy}
\pi^* = \operatorname*{argmin}_{a \in A_s}  Q(s,a)
\end{equation}

In our fleet scenario, the system dynamics are determined by drivers outside the system and fleet vehicles. This is modeled by the transition probabilities of the MDP. Given the policy of an agent, the choice of its actions can be determined. This information can be shared with other agents to reduce uncertainty about the system in the future by adapting the transition probabilities accordingly. Therefore, the MDP associated with each agent needs to contain the location and destination of all other agents.

We now formulate the fleet DRR problem as an MDP:
\begin{itemize}
	\item \textbf{State} $s=(l, \{L_i\}, \{ D_i \}, \{ r_j \})$, where $l \in N$ is the node at which the vehicle is currently located, $\{L_i\}$ is the set of the position of all other agents, $\{D_i\}$ denotes the set of final destinations of these agents and $r_j \in \{\text{available}, \text{occupied}\}$ is the occupation state of the $j^{\text{th}}$ resource.
	\item \textbf{Actions} 
	\begin{itemize}
		\item \textit{Take Road Action:} This action corresponds to taking an edge of the road network. Exactly one \textit{Take Road Action} exists for every outgoing edge from the current node. 
		\item \textit{Take Resource Action:} This action corresponds to driving to the resource, parking and walking to the final destination. Thus it always leads to the terminal state. There is one \textit{Take Resource Action} for every available resource located on an outgoing edge of the current node. 	\end{itemize}
	\item \textbf{Costs} $c(a)$, where $a \in A_s$ :
	\begin{itemize}
		\item \textit{Take Road Action:} The cost of this action consists of the travel time needed to traverse the road, i.e., the cost of the edge in the road network graph $G$.
		\item \textit{Take Resource Action:} The cost of this action consists of the time needed to drive from the intersection to the resource $r$ and the terminal cost $c_T(r)$, which is the time needed to park and walking to the final destination.
	\end{itemize}
	\item \textbf{Transition Probability} An action determines the position of the agent in the next state. However, the state of the resources are not certain. The probability $P(s' | a, s,  \{ \pi_i \}) $ of the next state $s'$ depends on the action $a$ and the static policies $\{\pi_i \}$ of the other agents:
	\begin{itemize}
		\item \textit{Take Road Action:} In this case, the transition probability is the product of the transition probabilities of all resources according to the probabilistic model, after the time for traversing the edge has passed. 
		\item \textit{Take Resource Action:} This action leads to the terminal state with probability 1, as the agent has parked.
	\end{itemize}
	If any of the policies of the other agents selects a \textit{Take Resource Action}, the resource transitions to the occupied state with certainty.
\end{itemize}

Finding an optimal policy for the fleet agents in the setting above is computationally very expensive and thus infeasible for larger sets of agents and resources. It has already been shown in  \cite{Schmoll:2019:SDR:3340964.3340983} that solving the DRR problem for a single agent is computationally problematic for larger settings. The fleet DRR now adds to this complexity. In single-agent settings, the state space complexity is exponential in the number of resources. In the multi-agent setting, the state space is additionally exponential in the number of agents. Consequently, we do not examine the search for an optimal policy but propose heuristics that improve the performance compared to single-agent solutions. Therefore, we will recapitulate the current state-of-the-art approximate solutions to single-agent DRR and compare our method to those. 

\subsection{Continuous-Time Markov Models}
Most methods in this paper require a stochastic process for describing the future availability of parking spots relative to the last time the parking spot was observed.
\cite{Josse2015} propose to use continuous-time Markov chains (CTMC) for modeling the time-dependent availability of each resource to incorporate short term observations (real-time availability through sensors) and long term observations (average vacancy/occupancy duration). The availability of each resource is given for the current time $t=0$. There exists a CTMC for each resource and the CTMCs of all resources are assumed to be mutually independent. A CTMC $X_i(t)$ with $t \geq 0$ can be defined by its state space $S = \{a, o\}$, indicating the availability of a parking spot, an initial probability distribution $\rho_0$ and a generator matrix $Q$. If real-time resource occupancy information is available, the initial probability distribution equals $\rho_0 = (1, 0)$ if the resource is available and $\rho_0 = (0, 1)$ if it is occupied. The generator matrix $Q^{dim(S) \times dim(\rho_0 )}$ describes the rate the CTMC moves between states. The transition probabilities depend on the parameters of the random variables modeling the sojourn times, the time a resource stays available and the time it stays occupied.
 \begin{equation}
 Q =\begin{pmatrix}
 -\lambda & \lambda \\
 \mu & - \mu
 \end{pmatrix}
 \end{equation}
The diagonal entries ($-\lambda$ and $-\mu$) reflect the sojourn times for staying in the current state, while other entries denote the rate departing from the current state into another. Using the Kolmogorov equations, we can compute the transition matrix from $Q$ by solving the differential equation $P'(t) = P(t)Q$, which leads to the following transition probabilities:
 \begin{equation}
 P(t) =	\begin{pmatrix}
 T^t_{a,a} = \frac{\mu}{\lambda + \mu} + \frac{\lambda}{\lambda + \mu} e^{-(\lambda + \mu)t} 
 & T^t_{a, o} = 1- T^t_{a,a}  \\
 T^t_{o,a} = \frac{\mu}{\lambda + \mu} - \frac{\mu}{\lambda + \mu} e^{-(\lambda + \mu)t}
 & T^t_{o,o} = 1- T^t_{o,a}
 \end{pmatrix}
 \end{equation}
 We denote $T^t_{f, f'}$ as the probability for being in state $f'$ after time $t$ has passed while in state $f$ at $t=0$. As aforementioned, the sojourn times are modeled as exponentially distributed random variables with parameter $\lambda$ when in state $available$ and $\mu$ when $occupied$. Therefore, $\lambda^{-1}$ describes the average time a resource stays available and $\mu^{-1}$ the average time a resource stays occupied.

\subsection{Replanning}
\label{sec:replanning}
While exactly solving an MDP tends to be slow, deterministic planning is often faster due to the use of intelligent heuristics in order to only examine a small part of the state space \cite{Little2007}. Replanning approximates a solution for a probabilistic planning problem by determinizing the most likely future and solving the determinized problem with a deterministic planner. When executing the solution and encountering an unexpected state, a new solution is computed from the current state. 

The replanning solution in \cite{Schmoll:2019:SDR:3340964.3340983} avoids working on the state space of the MDP but instead utilizes an extended street network for deterministic planning. Therefore, the network graph $G$ is extended by adding a virtual goal node $N_{\text{goal}}$ to model the \textit{Take Resource Action}, which represents claiming the resource and walking to the final destination. For each resource $r_i$, a virtual edge from any intersection $N_i$ to $N_{\text{goal}}$ is added, where $N_i$ is the starting node of an edge $r_i$ is assigned to. The cost $c_v(r_i) = c(N_i, N_{\text{goal}})$ of this virtual edge is set to the time needed to drive along the road to reach this resource and the terminal cost $c_t$. If the resource is not available, the cost is set to the expected time required to drive around the block until the resource becomes available again. The time for such a trip around a resource $r_i$ is denoted as $t_{\text{tr}}(r_i)$. 
Because the planning is not done in the state space of the MDP, we cannot determine the exact transition probabilities between the resource states.  However, this is not necessary in many DRR problems because the time required to travel from intersection to intersection is typically rather small compared to the average time between resource state changes. Thus, for the most likely future it can be assumed that a resource does not change its state. In a majority of settings, the replanner can detect mistakes very early if the target resource changes its state and thus, it does not suffer a large penalty if it has to replan a route.
In other words, since a non-terminal action (proceeding to the next edge) requires relatively small amounts of time, DRR often can be considered as \enquote{probabilistically uninteresting} \cite{Little2007} and replanning is a suitable approximate solver for DRR.

\subsection{Hindsight Planning}
\label{sec:hindsight}

Replanning has some drawbacks, especially in \enquote{probabilistically interesting} tasks \cite{Schmoll:2019:SDR:3340964.3340983, Yoon2008}. 
The idea behind hindsight planning is to approximate the value of a state by sampling a set of deterministic futures settings, optimizing these with a deterministic solver in hindsight and then, combining the solutions. This is often computationally less expensive than solving the probabilistic problem itself. 
\cite{Schmoll:2019:SDR:3340964.3340983} proposed a hindsight planner for DRR. Creating a determinization or future~$D$ is done by determinizing the state of resources in the future. Let $C(s, a^*, D)$ denote the costs of the optimal solution~$a^*$ in state~$s$ using the determinization~$D$. The hindsight utility value or expected costs is defined as follows:
\begin{equation}
U_{\text{hs}}(s) = E_D[C(s', a^*,D)]
\end{equation}
Using this definition, we can now define $Q_{\text{hs}}(s, a)$:
\begin{equation}
\hat{Q}_{\text{hs}}(s, a) = c(a) + E[U_{\text{hs}}(s')]
\end{equation}
An optimal policy $\pi^*$ can be approximated by taking the action which has the best expected one-step look ahead hindsight value:
\begin{equation}
\pi^*(s) \approx \operatorname*{argmin}_{a \in A_s} \hat{Q}_{\text{hs}}(s, a)
\end{equation}
The policy $\hat{Q}_{\text{hs}}$ approximates the $Q$-function by exchanging the order of minimization and expectation, i.e., we no longer take the policy with minimum expected cost but use the expected cost of optimal policies w.r.t. $D$. This can be efficiently approximated by solving determinizations of the probabilistic problem. Note that $U_{\text{hs}}$ is a lower bound of the optimal utility values $U^*$ as it is assumed that the outcome of each action is known.

In DRR, sampling a future is essentially equal to sampling from the probability distribution of the resource states. The optimal solution of that determinization is choosing an upon arrival available resource with the least total costs (which includes the costs of driving to the resource and walking to the final destination). The hindsight costs $C(s, a, D)$ can be computed by first calculating the arrival time at each resource from the current position. In the next step, a determinization of the resource state at arrival time is created by sampling using the availability probability from the prediction model. If the resource is by any chance not available, we need to sample the time needed until the resource becomes available again. As a driver is not allowed to wait in front of a resource, we need to sample the number of round trips around the block until that resource becomes available again. However, sampling these is very time-consuming. Therefore, sampling is replaced with the mean time until a resource becomes available again, denoted as the minimum expected wait time $t_{\text{claim}}$.

\section{Reservations}
\label{sec:reservations}
In our multi-agent setting, a resource can become occupied either through a fleet vehicle or through a vehicle which is not part of the fleet. The latter case cannot be prevented and is anticipated by a model for resource state changes, in our case a continuous-time Markov chain (CTMC) \cite{Josse2015, Schmoll:2019:SDR:3340964.3340983}. To reduce uncertainty, intended occupations by other fleet vehicles need to be considered. This is motivated by the fact that any other fleet vehicle heading for the same parking spot and arriving earlier than the agent dramatically reduces the probability that the spot is available when the agent arrives. We call this process reservation as the spot can be considered as occupied even if it is currently not. 
More formally, a reservation is a tuple $res=(r_i, \alpha_i, t_{\text{arrival}})$, where $t_{\text{arrival}}$ denotes the time when $\alpha_i$ arrives at $r_i$. Let us note that a resource with a reservation still can be occupied by drivers not controlled through the system or by agents from the fleet arriving at the resource before the reservation becomes active.
\subsection{Replanning with Reservations}
\label{sec:res_replanner}
Replanning chooses the route to the resource with the least total costs. In this approach, the arrival time of an agent at its target resource is known. Therefore, other agents can treat this resource as if it was occupied in case they expect to arrive later. If an agent changes the target resource, an existing reservation will be deleted and a reservation for the new target resource will be created. Every time the availability of a resource is checked, one also needs to take the arrival time at that resource into consideration.
In \cite{Schmoll:2019:SDR:3340964.3340983}, the replanner can efficiently replan by using the efficient D*-Lite algorithm. With reservations, the application of this algorithm is not possible because the costs of virtual edges depend on the availability of a resource. However, the availability of a resource depends on the arrival time and therefore, on the path previously taken. Finding the shortest path in a time-dependent graph is a NP-hard problem in general \cite{Dean2004,Foschini2011}. Pre-computing all pairwise travel times between intersections can be done in polynomial time \cite{10.1145/367766.368168,Pettie2004} as we do not consider time-dependent costs. Calculating the arrival time at each resource can be achieved in constant time during the query phase. The memory cost can be reduced to $O(N^2)$ \cite{10.1145/1376616.1376623}.

\subsection{Hindsight Planner with Reservations}
\label{sec:res_hindsight}
A hindsight planning agent does not target a specific resource until the very end when a \textit{Take Resource Action} is chosen. To generate reservations, we compute the most often visited resource in all determinizations and the expected arrival time at that resource. Based on this estimate, we can generate a reservations for each agent. An exception are agents choosing the \textit{Take Resource Action}. For these agents, a very short-term reservation is created instead. This reservation resolves conflicts in the very near future when the parking intention of an agent is certain. Reservations can now be used to restrict sampling of the possible futures, as all reserved resources can be considered as taken in all possible futures.

\section{Multi-Agent Dynamic Probability Adaption}
\label{sec:prob_adaption}
The hindsight planner with reservation overestimates the probability that a resource is available because it does not incorporate the behavior of other agents when they fail to take their most often visited resource. To better reflect these situations, we propose an algorithm in which the behavior of an agent, in case the most often visited resource is not available, is approximated efficiently through biased random walk on the road network.
Availability probabilities of resources around the most often visited resource are decreased, according to the outcome of the simulated behavior.
In contrast to a pure random walk, the jumping probabilities of a biased random walk are not equal and can depend on several factors, including the previously chosen nodes \cite{Azar1996}. Such a random walk is well suited to efficiently approximate the aforementioned behavior because of its structural similarity to the DRR MDP. At each intersection in the MDP, an action needs to be selected: Whether to drive to another intersection or to park at a reachable resource. The policy specifies a probability distribution over the set of those actions. The biased random walk aims to estimate these probabilities to approximate the policy efficiently. 

In order to calculate the probability of taking another resource, we propose a self-interacting time-dependent biased random walk on a subset of the graph within an isochrone around the targeted resource to limit the number of streets that need to be taken into consideration. Let $N_i$ be the node where the agent is currently located and $N_j$ is a node reachable from $N_i$. The set of resources located on the edge $(N_i, N_j)$  is denoted as $R_{\text{reachable}}(N_i,  N_j)$. The probability $P_{i,j}(t) $  that at least one resource on the edge $(N_i, N_j)$  is vacant at time $t$ is defined as follows:
\begin{equation}
P_{i,j}(t) = 1-\prod_{r \in R_{\text{reachable}}(N_i, N_j)} P(r=\text{occupied}, t)
\end{equation}
Note that we assume that the availability probabilities of the resources are mutually independent. We calculate the bias  $\gamma_{i,j} \in ]0,1]$ for our random walk as the product of visit decay factor~$\theta_{i,j}$ and a penalty factor~$\delta_{i,j}$ for moving further away from the target:
\begin{itemize}
	\item $\theta_{i,j} = \begin{cases}
	0.95 &  \textrm{if edge i,j  was already visited}\\
	1 & \textrm{otherwise}
	\end{cases}$
	\item $\delta_{i,j} =  \frac{c((N_i,N_j), \text{target}(\alpha))}{\text{IsochroneLimit}}$, with $c((n_i,n_j), \text{target}(\alpha))$ being the time driving from the end of the road to the final destination of the agent $\alpha$.
\end{itemize}
The jumping probability is defined as follows:
\begin{equation}
J_{i,j}(t) = \gamma_{i.j}  (1- \prod_{r \in R_{i,j}} P_{i,j}(r=\text{occupied}, t))
\end{equation}
With each jump from $n_i$ to $n_j$, the accumulated time $t_{\text{acc}}$ is increased by the costs of driving $c(N_i, N_j)$. The random walks starts at the end of the road in which the target resource is located because agents can only make decisions when they are at intersections. The accumulated time $t_{\text{acc}}$ is initialized with the time needed to drive from the target resource to the next intersection. The path probability $P_{\text{path}}$ reflects the likelihood of not having found a resource and being at the end of the path. This probability is the product of all jumping probabilities $J_{i,j}$ for the path. It is initialized with the probability $P_{\text{initial}}$ that the preferred resource is occupied. We then sample how often we end up at a resource $r$ by only considering resources located on the last street of the random walk. A random walk may end after each jump with probability $1-P_{\text{path}}$. Let $E_{t_{\text{arrival}}}(r)$ denote the expected arrival time at a resource. It can be calculated using the mean accumulated times $t_{\text{acc}}$ of paths that end in a street from which $r$ is reachable. Probability adaptions for a resource are applied by subtracting the parking probability $P_{\text{park}}(r)$ from all predicted availability probabilities after time $E_{t_{\text{arrival}}}(r)$. The parking probability $P_{\text{park}}(r)$ is calculated by equally distributing the expected path probabilities $E[P_{\text{path}}]$ to all resources located on that street. When an agent changes its target resource, all adaptions created by that agent are reversed and the process is repeated with the new target resource.
Note that using the expected arrival time at the resource $E_{t_{\text{arrival}}}(r)$ is not entirely accurate. However, it increases the computational performance. As a further enhancement, one could create multiple adaptions for each walk in order to create more precise predictions about when agents arrive at certain parking spots.
The algorithm for creating probability adaptions is described in Algorithm \ref{alg:prob_adap} and Algorithm \ref{alg:path_to_adaption}.

\begin{algorithm}[t]
	\caption{Perform the biased random walk in order to create probability adaptions for resource $r_{\text{target}}$ with an expected arrival time $t_{\text{arrival}}$.}\label{alg:prob_adap}
	\begin{algorithmic}[1]
	    \small
		\Function{AdaptProbabilities}{$r_{\text{target}}, t_{\text{arrival}}$}
		\State $paths$ = $\emptyset$
		\For{i = 0 $\ldots$ $ samples $}
		\State $P_{\text{path}}$ = $P(r_{\text{target}}=\text{available}, t_{\text{arrival}})  $
		\LineComment{initial probability for performing a specific walk}
		\State $currentNode$ = $r_{\text{target}}.\text{road}.\text{to}$
		\State $t_{\text{acc}}$ = $t_{\text{arrival}} + t_{\text{partial}}$
		\While{true}
		
		\State $\vec{p}$ = calculate biased probability $J_{i,j}(t_{\text{acc}})$ of each \phantom  o \phantom  . \phantom .  \phantom .  \phantom  . \phantom . \phantom . \phantom . \phantom .  outgoing edge
		
		\State $\vec{p'} = \frac{\vec{p}}{\sum_{p_i \in \vec{p}} p_i} $
		\LineComment{normalize probability vector}
		
		\State $\vec{p'}_k = \sum_{i=0}^{k} \vec{p'}_i$
		\LineComment{cumulative sum of probability vector}  
		
		\State $\vec{p'}_k = \sum_{i=1}^{k} (\vec{p'}_{i-1}, \vec{p'}_i)$
		\LineComment{create probability vector interval}  
		
		\State $random \in U[0,1]$
		\LineComment{choose from uniform distribution}
		\State $nextEdge$ = choose edge $k$ with $random \in p_k$
		\\
		\State $P_{\text{path}}$ *= $\vec{p}_k$
		\State $t_{\text{acc}}$ += $c(nextEdge)$
		\State $currentNode$ = $nextEdge$.road.to
		\\
		\If{$random > p_k$ or no edges available}
		\State  $paths$.add(($currentPath$, $pathProbability$, \newline
        \hspace*{5.9em} $arrivalTime$))
		\State break
		\Else
		\State $currentPath$.add(($nextEdge$)
		\EndIf
		\EndWhile
		\EndFor
		\State CreateAdaptions($paths$)
		\EndFunction
	\end{algorithmic}
\end{algorithm}

\begin{algorithm}[t]
	\caption{Create Adaptions from the set of paths.}\label{alg:path_to_adaption}
	\begin{algorithmic}[1]
	    \small
		\Function{CreateAdaptions}{$paths$}
		\State $groupedPaths$ = group $paths$ by last road
		\ForAll{$road$, $p'$ = [($path$, $P_{path}$, $t_{\text{arrival}}$)] $\in$ $groupedPaths$}
		\State $E[t_{\text{arrival}}] = \frac{\sum_{t_{\text{arrival}} \in p'} t_{\text{arrival}}}{\lvert p' \rvert}$
		\\
		\State $E[P_{path}] = \frac{\sum_{P_{path} \in p'} P_{path}}{\lvert p' \rvert}$
		\\
		\State $R_{road}$  = resources reachable from $road$
		
		\ForAll{$r \in R_{road}$}
		\State $P_{\text{park}}(r) = \frac{E[P_{path}]}{\lvert  R_{road} \rvert}$
		\State UpdateProbabilities($P_{\text{park}}(r)$, $E[t_{\text{arrival}}]$)
		\LineComment{Store and update availability probabilities \newline
        \hspace*{5em} accordingly}
		\EndFor 
		\EndFor
		\EndFunction
	\end{algorithmic}
\end{algorithm}

\section{Evaluation}
\label{section:simulator}

In this section, we will describe how we evaluated our approaches, present the results and discuss them. The experiments have been conducted through an agent-based simulation. All algorithms have been implemented in Java and the simulation has been executed on a Linux server VM with a single thread for each run. The system featured an Intel(R) Xeon(R) Silver 4108 CPU running at 1.8~GHz and 59~GB of RAM. The least-cost paths between all pairs of intersections and the \textit{Minimum Expected Wait Time} have been pre-computed. Our agent-based simulation is based on a modified version of the COMSET simulator created by \cite{COMSET}. Let us note that even though COMSET was built for a taxi assignment task, it can be adapted to parking search by redefining passengers to parking spots and changing the pickup action to a take resource action.

\subsection{Setting}
\subsubsection{Road Network and Resources}
\begin{figure}[t]
	\centering
	\includegraphics[width=\linewidth]{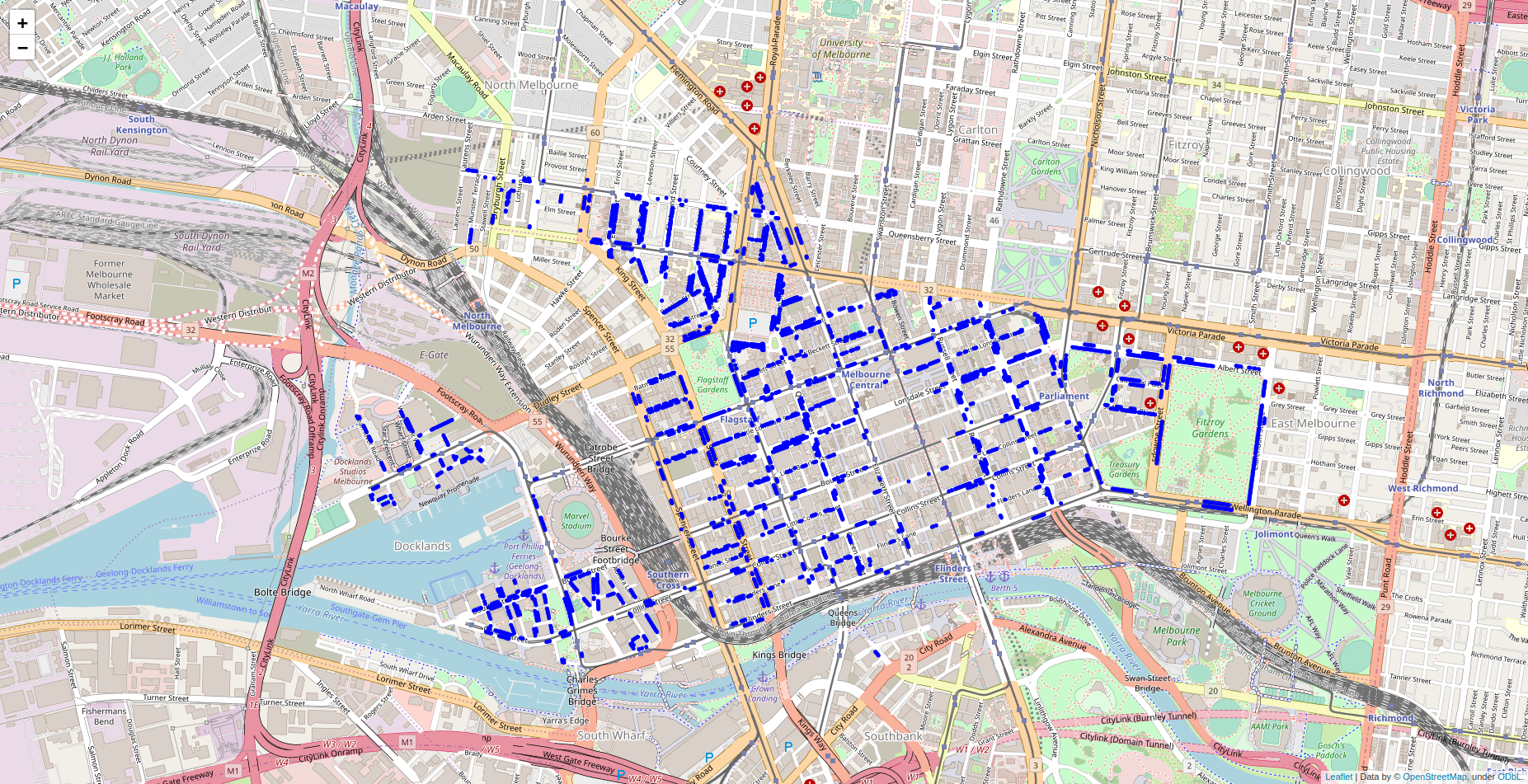}
	\caption{Locations of parking spots present in the simulation visualized as blue dots.}
	\label{fig:all_parking_spots}
\end{figure}

All experiments were conducted on a sub-graph of the Melbourne road network consisting of 3185~nodes and 6384~edges obtained from OpenStreetMap. The resulting road network has a total length of 305~km and an average road segment length of 48~meters. Attached to the edges are 4608~on-street parking spots, also shown in Figure \ref{fig:all_parking_spots}, whose locations are published by the City of Melbourne under the creative commons license \footnote{On-street parking spot dataset: https://data.melbourne.vic.gov.au/Transport-Movement/On-street-Parking-spots/crvt-b4kt}.
In order to accommodate for traffic jams and turn delays, the driving speed of an agent is set to 25\% of the road segment's speed limit. This factor has been calibrated through taxi trip times in Manhattan recorded in the TLC Yellow dataset. The speed limits are acquired from OpenStreetMap. 
The terminal costs $c_T$ of a resource are defined as the time an agent needs to walk from the resource to its final destination. The distance from the resource to the final destination is measured by greater circle distance. This means an agent can walk through buildings or perform arbitrary road crossings. The walking speed is set to 1.42 meters per second according to \cite{doi:10.1152/japplphysiol.00767.2005}.

\subsubsection{Resource Volatility}
How long a parking spot stays vacant and how long it is occupied has a huge influence on the difficulty of finding a parking spot. To evaluate the algorithms in various environments, we use a real-world setting with data from in-ground parking sensors and a synthetic setting.
\paragraph{Real-World Resource Occupations}
Real-world occupation data have been recorded by the city of Melbourne through in-ground sensors \footnote{Occupation dataset: https://data.melbourne.vic.gov.au/Transport/On-street-Car-Parking-Sensor-Data-2017/u9sa-j86i}. The temporal resolution is one second, which allows realistic replay. A two hour time period on a working day in 2017 is used in our simulation.
The empiric average time a resource stays available is approx. 75~min; the average time it stays occupied approx. 29~min.
\paragraph{Synthetic Resource Occupations}
To determine the influence of the prediction model, we also created synthetic resource occupations by sampling sojourn times from the exponential distributions of our CTMC model. We use the CTMC parameters $\mu^{-1} =$ 2091~seconds and $\lambda^{-1} =$ 120~seconds, i.e. parking spots are assumed to be occupied for about 35~min on average and occupied again after an expected time of 2~min of being free. We believe that this setting resembles situations where vacant parking spots are rare (an expected proportion of 5.4\% of the resources are vacant at any given time) and as a result, drivers have to compete over available parking spots.
Compared to the real-world setting described above, the synthetic setting has a higher resource volatility as well as a higher occupation percentage and thus presents a greater challenge to parking search algorithms.

\subsubsection{Agents' Final Destinations}
\begin{figure}[t]
	\centering
	\includegraphics[width=\linewidth]{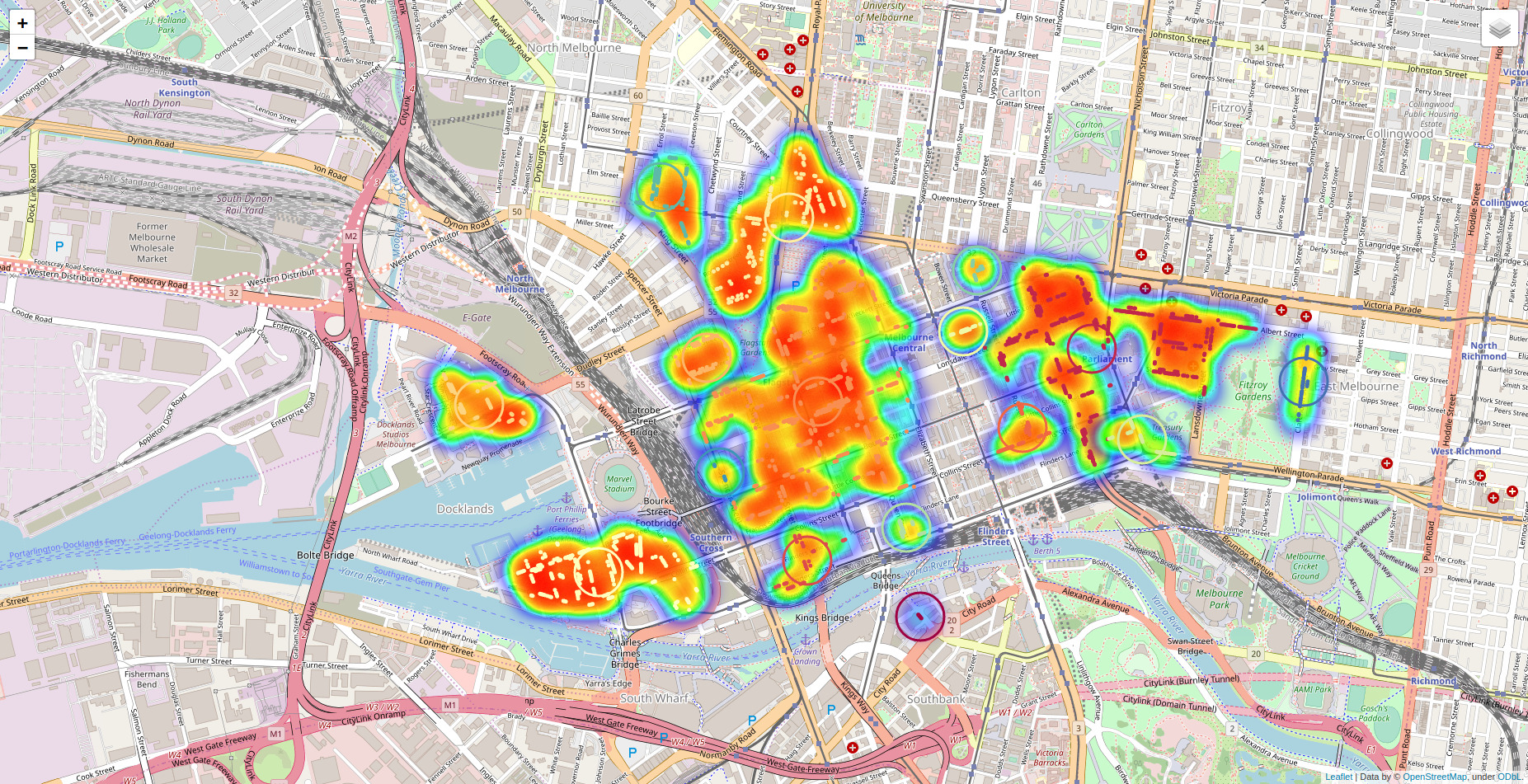}
	\caption{A heatmap of the parking occupations in different clusters during one hour on a working day.}
	\label{fig:cluster_map}
\end{figure}

\begin{figure}[!tb]
	\centering
	\includegraphics[width=\linewidth]{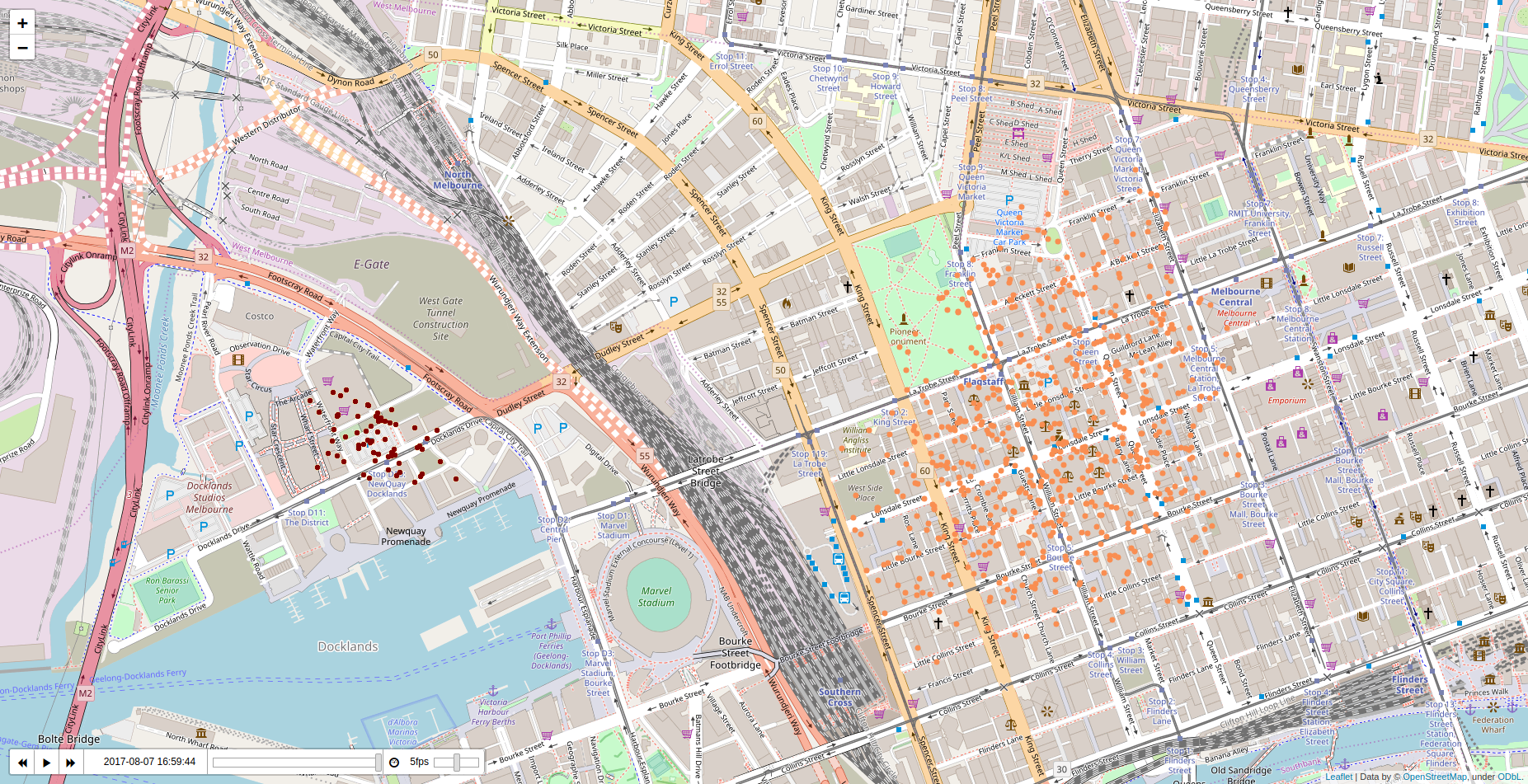}
	\caption{Destinations of agents chosen from empirical data in two clusters with very high (center right, orange dots) and high demand (center left, dark red dots) for parking.}
	\label{fig:cluster_agents_destinations}
\end{figure}

As motivated in the introduction, settings in which agents compete over resources are especially interesting. We used two settings for the agents' destinations in this paper, one high-competition setting in which the competition effects can be evaluated especially well and another setting in which destinations have been derived from empirical data. In both, agents start at the same point in the northeast corner of the map. An agent finished its trip when it has parked at a resource and walked to the final destination.
\paragraph{Single Destination}
In this setting, we artificially create a high competition over resources by placing 20 agents at the same time on the map with the exact same destination. This setting is very well suited to examine how different algorithms can handle competition over resources.
\paragraph{Data-Driven Destinations}
In order to get close to the real world, we aim to approximate both destinations and the number of drivers looking for a parking spot from real-world parking occupation data. As the targets of agents depend on the time of the day, we calculate the destinations on an hourly basis. For every occupation of a parking spot during that hour, a single point is created. In the next step, DBSCAN \cite{Ester:1996:DAD:3001460.3001507} is applied to infer clusters from those points. A visualization of these clusters is shown in Figure~\ref{fig:cluster_map}. The number of points inside the cluster corresponds to the number of agents heading towards a destination inside the cluster.
We carried out the evaluation during two hours of a working day. The times when agents start their trip have been randomly selected and are uniformly distributed over a period of one hour. We chose two clusters and randomly selected destinations inside these clusters, as shown in Figure~\ref{fig:cluster_agents_destinations}. 729~agents are heading into the first cluster while 63~agents have their final destination in the other cluster. The total number of agents is~792. Note that agents have varying trip times due to their varying destinations.

\subsubsection{Approaches}
We evaluated a number of approaches, which have been implemented as individual agents in our simulation. 
The implemented replanning apporach is the spatial replanner with improved cost model \cite{Schmoll:2019:SDR:3340964.3340983}.
Hindsight planning based approaches used 100 determinizations to sample the resource states.
Hindsight planning  with reservations is based on the resource that has been chosen in most determinizations.
The probability adaptions have been created by performing 30 biased random walks within an isochrone of 300 seconds driving time around the resource that has been chosen in most determinizations. 
We also included two baseline approaches not using availability information: Firstly, an agent which drives to the destination street and then continues taking random streets until an available parking spot is found. Secondly, we included Parkagent\cite{Benenson2008} with its default parameters in order to simulate the behavior of a human driver.

\subsection{Metrics}
Several metrics were used in our evaluation:
\paragraph*{Total Trip Time}
The time an agent needed for driving from start to an available parking spot plus the time spent walking from there to the final destination.
\paragraph*{Parking Time}
Comparing total trip times is not suitable for measuring the overhead caused by the need to park when the trip length of agents varies because of differing origins or destinations. Thus we introduce taxi time as the time an agent would need when taking a taxi instead (assuming the taxi is immediately available). A taxi does not need to park and drops off the passenger as close as possible to the final destination. The parking time is the difference between total trip time and taxi time, i.e., the actual overhead of parking search we are typically interested in.
\paragraph*{Number of Unsuccessful Resource Claims}
When an agent fails to take the resource it targets because it is occupied at arrival, we call this an unsuccessful resource claim.
\paragraph*{Computation Time}
The computation time is the time an agent spent in order to plan routes.

\subsection{Results}
We now present the results of our agent-based simulation. First, we cover the effectiveness of multi-agent improvements, i.e., the influence of sharing parking intention, and the overall performance of the respective agent implementations. We then compare the runtime efficiency of all evaluated methods.
Agents which did not find a parking spot within the simulation period are included with a total trip time of two hours.
In the following plots and tables, we refer to replanning algorithms as \textbf{RPL} and to hindsight planning algorithms as \textbf{HS}. Variations of these algorithms making use of reservations (see section~\ref{sec:reservations}) are denoted with suffix \enquote{\textbf{+~R}}; variations with probability adaptions (see section~\ref{sec:prob_adaption}) with suffix \enquote{\textbf{+~A}}.
\subsubsection{Effectiveness of Multi-Agent Improvements}
\begin{table}[!t]
\centering
\begin{tabular}{lllll}  
\toprule
& \multicolumn{2}{c}{Single Destination} & \multicolumn{2}{c}{Data-Driven Destinations} \\
\cmidrule(r){2-3}
\cmidrule(r){4-5}
\textbf{Algorithm}  & Real & Synthetic & Real & Synthetic \\
\midrule
\textbf{RANDOM} & 1634  & 842       & 794	    & 957   \\
\textbf{PARKAGENT} & 961 & 964      & 588	    & 555   \\
\midrule
\textbf{RPL} & 667      & 954       & 41	    & 104   \\
\textbf{HS}  &  515     & 476       & 122       & 95 \\
\midrule
\textbf{RPL + R} & \textbf{106}  & 276       &\textbf{34}	    & 100 \\
\textbf{HS + R}  & 109  &  262      & 97	    & \textbf{73} \\
\textbf{HS + A}  & 111  &\textbf{237}       & 37        & 81 \\
\bottomrule
\end{tabular}
	\caption{Mean parking time in seconds in fully observable settings.}
		\label{tb:full_taxi_offset_comp}
\end{table}

Table~\ref{tb:full_taxi_offset_comp} shows mean parking times of various approaches. In settings with real-world occupations, replanning with reservations delivers the best results, very closely followed by hindsight planning with adaptions. This is probably a result of using the CTMC, which is only an approximation of real-world parking spot behavior.
In synthetic occupation settings, hindsight planning approaches achieve smaller parking times than replanning approaches. In single destination settings with synthetic occupations, adaptions have the best parking times, while in data-driven settings reservations were slightly better.
All multi-agent approaches can provide effective parking guidance in various settings. Their parking time is always much lower than that of Parkagent, an agent designed to resemble a human searching for parking spots without occupancy information.

To evaluate the effectiveness of multi-agent improvements, we compare the relative difference in overall parking time of all agents with multi-agent improvements against their baseline approach. The results are detailed in Table~\ref{tb:full_rel_dif}.
Results demonstrate a significant reduction in parking times through our proposed multi-agent approaches in all settings. For example, in the data-driven destination setting with real-world resource occupations, adaptions reduced the total parking time by 75\% compared with single-agent hindsight planning. Overall, improvements lie between around 79\% and 14\%.
Hindsight planning with adaptions is on average more effective than hindsight planning with reservations in many settings.
Reservations combined with replanning have significantly reduced the parking time, especially in real-world occupation and single destination scenarios. However, in the data-driven destination setting with synthetic resource occupations, it only decreased the total parking time by around 4\%.

\begin{table}[t]
\centering
\begin{tabular}{lllll}  
\toprule
& \multicolumn{2}{c}{Single Destination} & \multicolumn{2}{c}{Data-Driven Destinations} \\
\cmidrule(r){2-3}
\cmidrule(r){4-5}
\textbf{Algorithm}  & Real & Synthetic & Real & Synthetic \\
\midrule
\textbf{RPL + R} & 84.11  &  71.11   &  18.11   & 3.91   \\
\textbf{HS + R}  &  78.85  & 44.85     & 22.69    & 22.93   \\
\textbf{HS + A}  & 78.38 & 50.15	&  75.17    & 14.44    \\
\bottomrule
\end{tabular}
	\caption{Reduction in percent of the total parking time by multi-agent improvements over their corresponding single-agent approach in various fully observable settings.}
	\label{tb:full_rel_dif}
\end{table}

\begin{figure}[t]
	\centering
	\includegraphics[width=\linewidth]{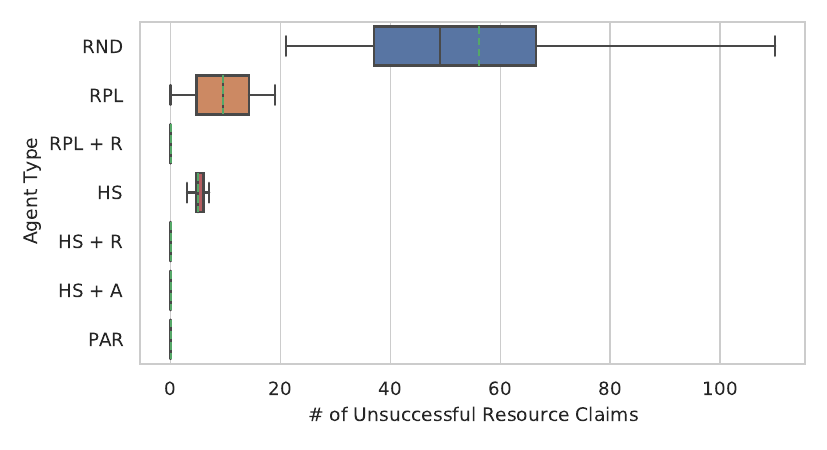}
	\caption{Number of unsuccessful resource claims in a fully observable single destination setting.}
	\label{fig:basic_full_real_unsuc_claim}
\end{figure}

In a multi-agent setting with real-world resource volatility and many agents heading to the same destination, the agents themselves have a large impact on the future states of resources. Multi-agent improvements proposed in this paper aim to incorporate the impact of the agents on future states of resources. An unsuccessful resource claim happens when an agent wants to park at a resource that has been occupied since it decided to park there. The time between the final decision to park at a resource and arriving at that resource is very short because agents make that decision at the intersection from which the resource is reachable. As this time frame is quite small, it is unlikely that the resource changes due to non-simulated parking events. Multi-agent improvements aim to reduce conflicts. Consequently, these approaches should have fewer unsuccessful resource claims. This assumption is confirmed by our experiments and is detailed in Figure~\ref{fig:basic_full_real_unsuc_claim}.

\subsubsection{Computational Effort}

\begin{figure}[t]
	\centering
	\includegraphics[width=\linewidth]{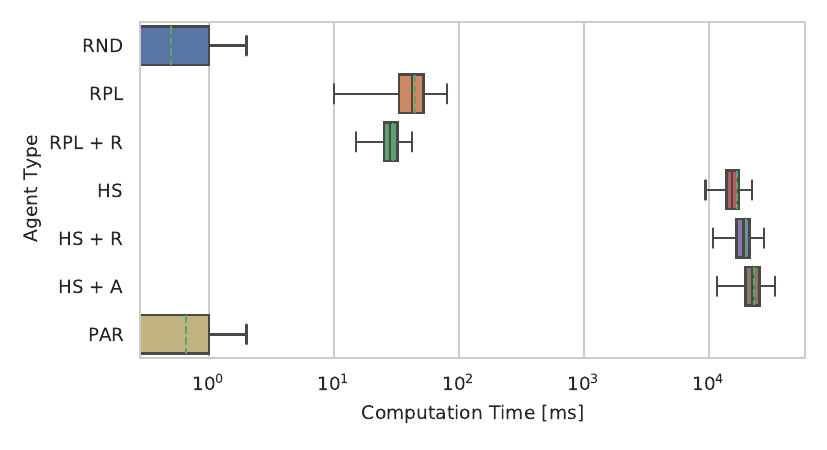}
	\caption{Computation time in ms on a logarithmic scale in a data-driven destination setting.}
	\label{fig:cluster_full_comp_time}
\end{figure}

Figure~\ref{fig:cluster_full_comp_time} compares the computation times in a data-driven destination setting. Other settings show very similar results, so we exclude these due to space restrictions. The computation time is the time spent planning for a whole trip. We observe that the median computation time of replanning based approaches is around 0.1~seconds, while those of hindsight planning based approaches is 10~s aggregated over the complete search trajectory. Reservations and adaptions do not have a significant influence on computation times. As our experiments have been conducted using a hundreds of agents and thousands of parking spots, these results lead to the conclusion that one can apply all approaches in real-time, even in large-scale scenarios. 

\section{Conclusion}
\label{sec:conclusion}
In this paper, we proposed multi-agent variations of existing approaches for solving the dynamic resource routing problem in fully observable scenarios: We formalized the problem and presented approximative approaches to solve it. These approaches, namely reservations and dynamic probability adaptions, solved each by a replanner and a hindsight planner, were evaluated through agent-based simulations in order to gain insights into the impact of sharing fleet data.
Given the results of our experimental evaluation, we have shown that both are able to significantly improve parking guidance. In situations with very few available parking spots, hindsight planning with reservations or adaptions can deliver the best results. Replanning with reservations does work very well in settings close to the real-world. It is efficient, easy to implement and can provide parking guidance without a prediction model. We come to the conclusion that all approaches presented in this paper can be deployed in real-world parking guidance systems of a vehicle fleet, given their effectiveness and efficiency, even in large scale scenarios.

The work presented in this paper concerns fully observable scenarios. However, not all cities are equipped with parking bay sensors, which makes partially observable settings (i.e. not all parking bay states are known) interesting. These can be approached with extensions to the methods given in this paper. We are currently working on these extensions and related experiments.

\section*{Acknowledgments}
We thank the City of Melbourne, Australia, for providing the parking datasets used in this paper under Creative Commons license.
This work has been funded by the German Federal Ministry of Education and Research (BMBF) under Grant No. 01IS18036A. The authors of this work take full responsibilities for its content.

\bibliographystyle{IEEEtran}
\balance
\bibliography{bib}

\end{document}